\PassOptionsToPackage{svgnames,table}{xcolor}
\documentclass[letterpaper]{article}
\usepackage[preprint]{tmlr}
\usepackage[T1]{fontenc}
\usepackage[utf8]{inputenc}
\usepackage{amsmath, url}
\usepackage{graphicx}
\usepackage{hyperref}
\usepackage{amssymb}
\usepackage{url}
\usepackage{enumitem}
\usepackage{float}
\usepackage{tikz}

\usepackage{amsfonts}
\usepackage{nicefrac}
\usepackage{microtype}
\usepackage{pifont}
\usepackage{tabularx}
\usepackage{array}
\usepackage{longtable}
\usepackage{tocloft}
\usepackage{pgfplotstable}
\pgfplotsset{compat=newest}
\usetikzlibrary{arrows.meta, positioning, fit, calc, shadows, shapes.geometric, chains, shapes.multipart, backgrounds, shapes, matrix, fit, patterns, arrows, decorations.pathreplacing, fadings, bayesnet, intersections, ext.node-families, ext.positioning-plus, ext.paths.ortho, automata, decorations.markings}

\title{\textbf{Engram Memory Encoding and Retrieval:\\A Neurocomputational Perspective}}
\date{}

\begin{document}

\author{\name Daniel Szelogowski \email dszelogowski@gmail.com \\
       \addr Department of Computer Science\\
       University of Wisconsin-Whitewater\\
       Whitewater, WI 53190, USA}

\maketitle

\vspace{-0.15cm}
\begin{abstract}
Despite substantial research into the biological basis of memory, the precise mechanisms by which experiences are encoded, stored, and retrieved in the brain remain incompletely understood. A growing body of evidence supports the engram theory, which posits that sparse populations of neurons undergo lasting physical and biochemical changes to support long-term memory. Yet, a comprehensive computational framework that integrates biological findings with mechanistic models remains elusive. This work synthesizes insights from cellular neuroscience and computational modeling to address key challenges in engram research: how engram neurons are identified and manipulated; how synaptic plasticity mechanisms contribute to stable memory traces; and how sparsity promotes efficient, interference-resistant representations. Relevant computational approaches --- such as sparse regularization, engram gating, and biologically inspired architectures like Sparse Distributed Memory and spiking neural networks --- are also examined. Together, these findings suggest that memory efficiency, capacity, and stability emerge from the interaction of plasticity and sparsity constraints. By integrating neurobiological and computational perspectives, this paper provides a comprehensive theoretical foundation for engram research and proposes a roadmap for future inquiry into the mechanisms underlying memory, with implications for the diagnosis and treatment of memory-related disorders.

\end{abstract}

\vspace{-0.2cm}
\section{Introduction}

\vspace{-0.25cm}
Memory, defined as the brain's capacity for storing and utilizing learned information, is indispensable for modulating behavior, enabling adaptation, and ensuring survival across species \citep{cite1}. The scientific quest to identify the physical substrate of memory has a rich history, culminating in the modern concept of the \textbf{engram}.
The notion that memories are stored as persistent alterations within the brain dates back to ancient Greek philosophers. However, its scientific articulation emerged in the early 20th century with Richard Semon. In 1904, Semon introduced the term ``engram,'' defining it as ``the enduring though primary latent modification in the irritable substance produced by a stimulus (from an experience)'' \citep{cite1, cite2}. He posited that an experience activates a specific population of neurons, which then undergo lasting chemical and/or physical changes to form an engram. Subsequent exposure to cues present during the original experience would then reactivate this engram, leading to memory retrieval --- a process Semon termed ``ecphory'' \citep{cite2}.

\vspace{-0.1cm}
While largely overlooked during his lifetime due to technological limitations, Semon's ideas proved remarkably prescient, foreshadowing many contemporary memory concepts, including the idea of a ``memory trace'' \citep{cite2}. His initial definition of an engram as a ``latent modification'' is particularly insightful. Modern discoveries of ``silent engrams'' --- which exist as physical traces but cannot be retrieved by natural cues, yet can be artificially reactivated --- directly align with Semon's concept of ``primarily latent modifications''. This congruence demonstrates that a robust theoretical abstraction, even without immediate empirical validation, can provide a guiding framework for scientific inquiry for over a century. The enduring quality of Semon's theoretical construct emphasizes the power of conceptual models in preparing for future experimental investigation. In the sections that follow, we build on this historical and biological foundation to propose a biologically grounded computational theory of engram encoding and retrieval. 

\subsection{Contemporary Understanding of Engrams/Engram Neurons}

Current neuroscience largely validates and expands upon Semon's foundational ideas. An engram is now understood as a group of neurons that become active during learning, undergo specific biochemical and physical alterations to store information in a stable state, and are subsequently reactivated during memory recall \citep{cite1}. These ``engram cells'' are considered the physical substrate where learning leaves lasting imprints within the brain. Memory formation involves three critical stages: \textbf{encoding}, where information from perception is written into the brain; \textbf{consolidation}, where this information is selected and stabilized for long-term storage; and \textbf{storage} itself.
A significant evolution in the understanding of engrams is the concept of an ``engram complex.'' This refers to the entire brain-wide representation of a memory, comprising functionally connected engram cell ensembles distributed across multiple brain regions \citep{cite2}. This aligns directly with Semon's original hypothesis of a ``unified engram-complex.''

\vspace{-0.2cm}
\begin{table}[h!]
\centering
\caption{Key Technologies for Engram Study}
\begin{longtable}{|>{\raggedright\arraybackslash}p{3.5cm}|>{\raggedright\arraybackslash}p{5.75cm}|>{\raggedright\arraybackslash}p{5.75cm}|}
\hline
\centering\arraybackslash\textbf{Technology} & \centering\arraybackslash\textbf{Mechanism/Application} & \centering\arraybackslash\textbf{Research Contribution} \\
\hline
\textbf{Optogenetics} & Expresses light-sensitive channels (e.g., ChR2) in engram neurons, allowing precise activation or inhibition with light. & Enables artificial memory recall, selective memory erasure, and creation of synthetic memories, demonstrating the necessity and sufficiency of engrams \citep{cite1, cite3}. \\
\hline
\textbf{Chemogenetics} & Utilizes Designer Receptors Exclusively Activated by Designer Drugs (DREADDs) for pharmacological control of engram neuron activity. & Offers sustained modulation of engram activity, similar to optogenetics but with slower kinetics \citep{cite1}. \\
\hline
\textbf{Immediate Early Gene (IEG) Labeling} & Hijacks promoters of IEGs (e.g., c-fos, Arc) to drive expression of fluorescent reporters (e.g., GFP) or functional channels in active neurons during learning. & Identifies and visualizes engram neurons; allows tracking of their activity and properties over time \citep{cite1, cite3}. \\
\hline
\textbf{GCaMP (Calcium Indicators)} & Enables \textit{in vivo} tracking of engram neuron activity by monitoring calcium fluctuations, which correlate with neuronal firing. & Provides real-time observation of engram neuron maturation and activity patterns during behavior \citep{cite3}. \\
\hline
\textbf{Electrophysiology} & Measures electrical activity (e.g., intrinsic excitability, synaptic currents) of engram neurons \textit{ex vivo} or \textit{in vivo}. & Characterizes physiological changes, such as increased excitability and synaptic inputs, in engram cells during consolidation \citep{cite1, cite3}. \\
\hline
\end{longtable}
\end{table}

\vspace{-0.2cm}
Modern technological advancements have revolutionized the study of engrams, enabling researchers to investigate how specific memories translate into neuronal changes with unprecedented resolution \citep{cite1}. These technologies include transgenic manipulation, optogenetics, chemogenetics, electrophysiology, and sophisticated behavioral techniques. \textbf{Immediate early genes (IEGs)}, such as c-fos and activity-regulated cytoskeleton-associated protein (Arc), serve as endogenous markers of neuronal activity. Researchers exploit their promoters to drive the expression of fluorescent reporters (e.g., GFP) or light-sensitive channels (e.g., channelrhodopsin-2, ChR2) in putative engram neurons, allowing for their identification and manipulation \citep{cite1, cite3}. Furthermore, the activity of engram neurons can be tracked in vivo during their maturation from encoding through consolidation using functional indicators like GCaMP (calcium indicators; \citealp{cite3}). These experimental manipulations, particularly in the hippocampus, have demonstrated the necessity and sufficiency of engram cells for memory functions, enabling selective memory erasure, artificial recall, and even the creation of synthetic memories.

The consolidation of an engram is not a static event but a dynamic process. Engrams undergo long-lasting synaptic modifications during this phase, and their properties evolve over time \citep{cite3}. For instance, CA3 pyramidal neurons recruited into an engram progressively acquire increased excitability over the first day following a one-trial memory task. This indicates that initial encoding triggers a dynamic consolidation phase involving intricate cellular and synaptic changes. This active refinement process enhances the retrievability and discrimination of memories, preventing different experiences from becoming indistinguishable \citep{cite4, cite20}. Such dynamic processes are critical for the brain's ability to form distinct and stable memory representations.

\subsection{Engrams in the Brain and Computational Neuroscience}

The study of engrams inherently bridges traditional neurobiology with computational neuroscience. The overarching goal is to translate fundamental findings from rodent engram studies into a comprehensive understanding of human memory acquisition, storage, and utilization, ultimately facilitating treatments for human memory and information-processing disorders \citep{cite2}.
Computational neuroscience employs mathematical tools and simulations to unravel the complexities of brain functions, including learning and memory, often conceptualizing the brain as a sophisticated computational system \citep{cite5, cite6}. Within this framework, engrams are often viewed as emergent behaviors in sparse neural networks, where specific groups of neurons remain active as memory traces following a learning event \citep{cite7}. The application of engram principles to \textbf{artificial neural networks (ANNs)} frequently involves architectural modifications, such as context-dependent gating mechanisms designed to allocate and manage distinct neuronal ensembles for different memories.

This field is characterized by a bidirectional influence between biological discovery and computational modeling. Biological observations of engrams directly inspire the development of novel computational models, as exemplified by the concept of ``engram gating'' in ANNs, which is explicitly derived from biological engram mechanisms \citep{cite7}. Conversely, computational studies can generate testable predictions about brain function that are subsequently investigated through biological experiments. For instance, initial computational models predicted a dynamic nature of engram populations, where the number of activated cells might decrease over time. Animal experiments later corroborated these predictions, which demonstrated that engram cells are dynamic and undergo refinement during consolidation \citep{cite4, cite9}. This reciprocal relationship reinforces computational neuroscience as a tool for simulating biological processes and a powerful framework for generating and refining hypotheses about the brain's intricate workings. 

\section{Engram Formation and Dynamics}

The precise biological mechanisms by which experiences are transformed into stable memory traces within the brain involve a complex interplay of cellular and synaptic modifications.

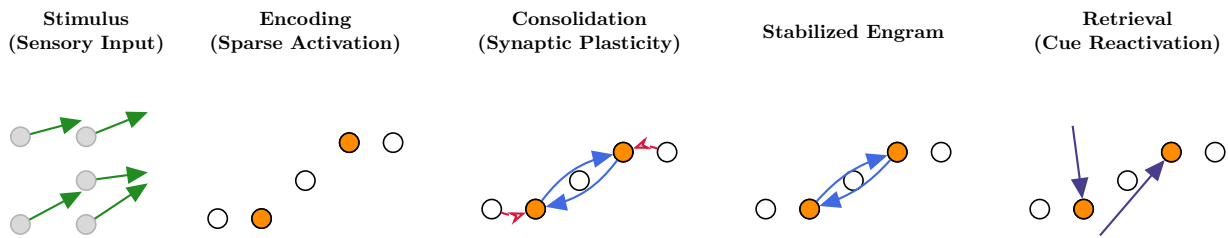
\begin{figure}[ht]
\centering
\begin{tikzpicture}[
    neuron/.style={circle, draw=black, fill=white, minimum size=10pt, inner sep=1pt},
    engram/.style={circle, draw=black, fill=DarkOrange, minimum size=10pt, inner sep=1pt},
    faded/.style={circle, draw=gray!60, fill=gray!30, minimum size=10pt, inner sep=1pt},
    excite/.style={->, thick, color=RoyalBlue},
    inhibit/.style={-{Stealth[open]}, thick, color=Crimson, dashed},
    stage/.style={font=\bfseries, text width=3.75cm, align=center},
    every path/.style={line width=0.6pt},
    scale=0.725, transform shape
]

\node[stage] at (0,2.7) {Stimulus\\(Sensory Input)};
\node[stage] at (4,2.7) {Encoding\\(Sparse Activation)};
\node[stage] at (9,2.7) {Consolidation\\(Synaptic Plasticity)};
\node[stage] at (14,2.7) {Stabilized Engram};
\node[stage] at (19,2.7) {Retrieval\\(Cue Reactivation)};

\node[faded] (s1) at (0,0) {};
\node[faded] (s2) at (-1.2,0.8) {};
\node[faded] (s3) at (0.0,0.8) {};
\node[faded] (s4) at (-1.2,-0.8) {};
\node[faded] (s5) at (0.0,-0.8) {};

\foreach \i in {1,...,5} {
    \draw[->, thick, color=ForestGreen] (s\i) -- ++(1.125,0.15*\i);
}

\foreach \i in {1,...,5} {
    \pgfmathsetmacro{\x}{4 + (\i-3)*0.8}
    \pgfmathsetmacro{\y}{0.8*sin(60*(\i-3))}
    \node[neuron] (e\i) at (\x,\y) {};
}
\foreach \i in {2,4} {
    \node[engram] (e\i) at ($(e\i)$) {};
}

\foreach \i in {1,...,5} {
    \pgfmathsetmacro{\x}{9 + (\i-3)*0.8}
    \pgfmathsetmacro{\y}{0.6*sin(60*(\i-3))}
    \node[neuron] (c\i) at (\x,\y) {};
}
\foreach \i in {2,4} {
    \node[engram] (c\i) at ($(c\i)$) {};
}

\draw[excite] (c2) to[bend left=20] (c4);
\draw[excite] (c4) to[bend left=20] (c2);

\draw[inhibit] (c1) to[bend right=20] (c2);
\draw[inhibit] (c5) to[bend right=20] (c4);

\foreach \i in {1,...,5} {
    \pgfmathsetmacro{\x}{14 + (\i-3)*0.8}
    \pgfmathsetmacro{\y}{0.6*sin(60*(\i-3))}
    \node[neuron] (s\i) at (\x,\y) {};
}
\foreach \i in {2,4} {
    \node[engram] (s\i) at ($(s\i)$) {};
}

\draw[excite] (s2) to[bend left=15] (s4);
\draw[excite] (s4) to[bend left=15] (s2);

\foreach \i in {1,...,5} {
    \pgfmathsetmacro{\x}{19 + (\i-3)*0.8}
    \pgfmathsetmacro{\y}{0.6*sin(60*(\i-3))}
    \node[neuron] (r\i) at (\x,\y) {};
}
\foreach \i in {2,4} {
    \node[engram] (r\i) at ($(r\i)$) {};
}
\draw[->, thick, color=DarkSlateBlue] (18,1) -- (r2);
\draw[->, thick, color=DarkSlateBlue] (18.5,-1) -- (r4);

\end{tikzpicture}
\caption{Lifecycle of an engram: An initial experience activates a sparse set of neurons, which undergo synaptic modifications during consolidation, forming a stabilized engram that can later be reactivated by partial cues. Excitatory (solid, blue) and inhibitory (dashed, red) connections guide engram refinement and stability.}
\label{fig:engram-lifecycle}
\end{figure}

\subsection{Cellular and Synaptic Changes in Engram Neurons}

Information derived from experiences is encoded in the brain as enduring changes within specific ensembles of cells --- i.e., engram cells --- which are essential for memory storage and recall \citep{cite10}. These cells represent a select subset of neurons that are activated by a particular experience and subsequently undergo molecular and structural modifications to form the memory trace. Their reactivation is both necessary and sufficient to elicit the associated behavioral response. %
During the encoding of novel information, multimodal sensory data converges onto brain regions such as the hippocampus, activating sparse neuronal assemblies. These assemblies are believed to form a memory representation through their concerted activity and strengthened synaptic interconnectivity \citep{cite3}. Critically, these nascent engrams then undergo long-lasting synaptic modifications during the consolidation phase and are selectively re-engaged when the memory is retrieved.%

Early investigations into these changes, specifically in CA3 pyramidal neurons, reveal significant alterations within hours of memory encoding. Studies using fast-labeling of engram neurons (FLEN) show that CA3 neurons recruited into an engram exhibit an increased number of excitatory inputs, evidenced by a higher frequency of spontaneous and miniature excitatory postsynaptic currents (sEPSCs and mEPSCs; \citealp{cite3}). This suggests a rapid increase in synaptic contacts or the efficiency of existing ones. While initial excitability may not differ, these CA3 engram neurons progressively acquire increased intrinsic excitability over the first day following learning, leading to more prolonged spiking activity and a greater number of action potentials. This heightened excitability likely enhances their readiness for future reactivation. Interestingly, an increase in inhibitory inputs, measured by miniature inhibitory postsynaptic currents (mIPSCs), is also observed in these engram CA3 neurons, both in the early and later phases of consolidation.

The simultaneous increase in both excitatory and inhibitory inputs to engram neurons during consolidation highlights a sophisticated interplay of plasticity mechanisms. The heightened excitation provides the necessary drive for memory encoding and retrieval, but without a compensatory increase in inhibition, this could lead to uncontrolled, runaway neuronal activity, compromising network stability. The increased inhibition likely serves to balance the heightened excitability, preventing pathological hyperexcitation and ensuring the precise and selective activation of engram ensembles. Hence, this intricate homeostatic regulation is crucial for maintaining network stability and specificity, allowing for the formation of distinct and well-defined memory traces.

\subsection{Synaptic Plasticity: The Molecular Basis of Engram Encoding}%

Synaptic plasticity, referring to the activity-dependent modification of synaptic strengths, is widely recognized as a fundamental physiological mechanism underlying learning and memory \citep{cite11, cite12}. The formation of an engram fundamentally involves the strengthening of synaptic connections among selected neurons during memory encoding, leading to the creation of ``synaptic engrams'' \citep{cite13}. This potentiation of synaptic strength is believed to increase the probability of recreating the same neural activity pattern during memory retrieval. Engram cells typically exhibit several hallmarks of synaptic strengthening, including a higher ratio of AMPA to NMDA receptors, an increased density of dendritic spines (which are sites of excitatory synaptic input), and an overall enhanced number of synaptic inputs \citep{cite10}.

At the molecular level, proteins such as \textbf{postsynaptic density protein 95 (PSD-95)} are critical in mediating these plastic changes. PSD-95 is a scaffolding protein located at the postsynaptic density, a specialized region of the neuron that receives synaptic input. It is crucial for synaptic function, influencing synaptic strength and maturation \citep{cite10}. Research indicates that PSD-95 impacts the specific connectivity patterns among engram cells and is vital for the long-term stability of memories. Dysregulation of PSD-95, as observed in knockout models, can lead to increased \textbf{long-term potentiation (LTP)} but also deficits in spatial learning, emphasizing its precise role in memory encoding and function. Synaptic engrams are thus characterized by ensembles of these strengthened synapses between engram neurons, encoding distinct memory traces \citep{cite13}.

\clearpage
Beyond these synaptic changes, learning and memory also involve ``intrinsic plasticity,'' which refers to growth processes or metabolic changes within the neuron itself, altering its excitability \citep{cite14}. Furthermore, non-synaptic plasticity, such as the regulation of neural membrane properties, can operate on faster timescales, potentially enabling rapid initial information storage, complementing the slower, more enduring synaptic plasticity processes \citep{cite12}. This suggests a multi-scale, coordinated plasticity mechanism at play in engram formation. Synaptic changes primarily encode the specific content of a memory by modifying the strength of connections between neurons, while intrinsic and non-synaptic changes modulate the overall responsiveness and participation of individual neurons within the engram. This coordinated interplay ensures both the precise encoding of information and the dynamic integration of neurons into stable memory circuits.

\subsection{Engram Consolidation, Stability, and Retrieval}

Following initial encoding, engrams undergo a critical consolidation phase characterized by long-lasting synaptic modifications. These consolidated engrams are then selectively re-engaged when the memory is retrieved \citep{cite3}. The activity of engram neurons can be tracked in vivo throughout their maturation, from the moment of encoding through the consolidation period. For efficient memory recall, the reactivation of engram cells typically needs to reach a certain threshold, often a percentage of the original engram population \citep{cite4, cite9}.
However, the engram population is not static. The number of engram cells activated during recall has been shown to decrease over time, indicating a dynamic refinement process. This dynamic nature serves a crucial function: during consolidation, the brain actively works to separate distinct experiences, effectively ``cutting away unnecessary neurons'' to refine and discriminate between different memories \citep{cite4}. This process is akin to refining a broad highway into distinct lanes, allowing for more precise memory access.

The plasticity of refinement extends to the phenomenon of forgetting. Natural forgetting is increasingly viewed not merely as a deficit of memory function but as an adaptive form of engram plasticity \citep{cite15}. It can represent a reversible suppression of engram ensembles, triggered by experience and perceptual feedback, which prompts cellular plasticity processes that adaptively modulate memory access, implying that the brain actively prunes or refines engrams to optimize memory storage and prevent information overload \citep{cite5}. This adaptive forgetting mechanism improves memory efficiency and discrimination by selectively reducing the accessibility or strength of less relevant engram components, ensuring that the most pertinent memories remain readily available and distinct.

\vspace{0.5cm}
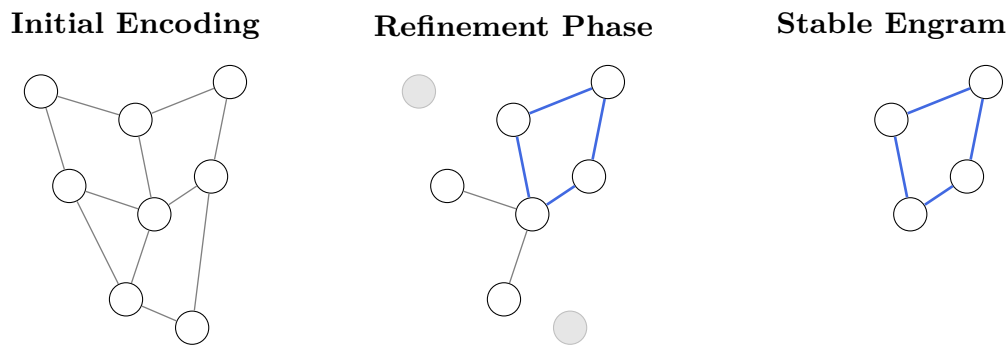
\begin{figure}[ht]
\centering
\begin{tikzpicture}[
    neuron/.style={circle, draw=black, fill=white, minimum size=10pt},
    strong/.style={line width=1pt, draw=RoyalBlue},
    weak/.style={line width=0.5pt, draw=gray},
    faded/.style={circle, draw=gray!50, fill=gray!20, minimum size=10pt},
    labelstyle/.style={font=\small\bfseries},
    scale = 1.25, transform shape
]

\node[labelstyle] at (1.5,4.2) {Initial Encoding};
\node[labelstyle] at (5.5,4.2) {Refinement Phase};
\node[labelstyle] at (9.5,4.2) {Stable Engram};

\foreach \name/\x/\y in {
  A1/0.5/3.5, B1/1.5/3.2, C1/2.5/3.6,
  D1/0.8/2.5, E1/1.7/2.2, F1/2.3/2.6,
  G1/1.4/1.3, H1/2.1/1.0%
}{
  \node[neuron] (\name) at (\x,\y) {};
}

\foreach \a/\b in {%
  A1/B1, A1/D1, B1/C1, B1/E1, C1/F1,%
  D1/E1, D1/G1, E1/F1, E1/G1, F1/H1,%
  G1/H1%
}{%
  \draw[weak] (\a) -- (\b);
}

\foreach \name/\x/\y in {%
  B2/5.5/3.2, C2/6.5/3.6,
  D2/4.8/2.5, E2/5.7/2.2, F2/6.3/2.6,
  G2/5.4/1.3
}{
  \node[neuron] (\name) at (\x,\y) {};
}

\node[faded] (A2) at (4.5,3.5) {};
\node[faded] (H2) at (6.1,1.0) {};

\foreach \a/\b/\style in {%
  B2/C2/strong, B2/E2/strong,
  D2/E2/weak, E2/F2/strong, G2/E2/weak, F2/C2/strong%
}{
  \draw[\style] (\a) -- (\b);
}

\foreach \name/\x/\y in {
  B3/9.5/3.2, C3/10.5/3.6,
  E3/9.7/2.2, F3/10.3/2.6%
}{
  \node[neuron] (\name) at (\x,\y) {};
}

\foreach \a/\b in {
  B3/C3, B3/E3, E3/F3, F3/C3%
}{
  \draw[strong] (\a) -- (\b);
}

\end{tikzpicture}
\caption{Structural plasticity during engram consolidation. Initially, many neurons are weakly connected after encoding. During refinement, unused neurons are pruned and relevant connections are strengthened. The final engram is sparse and robust, with strong internal connectivity supporting memory recall.}
\label{fig:engram-refinement-final}
\end{figure}

\section{Hebbian Plasticity: Learning and Memory}

Donald Hebb's seminal work in 1949 laid a theoretical foundation for understanding how neuronal activity could lead to learning and memory. His theory proposed a link between neurophysiological changes and cognitive processes through the formation of ``cell assemblies'' and ``phase sequences'' \citep{cite39}.

\subsection{Hebb's Postulate and the Formation of Cell Assemblies}

Hebb's neurophysiological postulate, famously summarized as ``\textit{neurons that fire together wire together},'' states: ``When an axon of cell A is near enough to excite a cell B and repeatedly or persistently takes part in firing it, some growth process or metabolic change takes place in one or both cells such that A’s efficiency, as one of the cells firing B, is increased'' \citep{cite14}. This postulate implies that learning involves not only changes at the synapse (synaptic plasticity) but also broader ``growth processes or metabolic changes'' within the neuron itself (intrinsic plasticity).

A ``cell assembly'' is a theoretical construct representing a set of neurons and their interconnected pathways that function as a cohesive unit. When a strong stimulus activates a particular cell assembly, it drives high-frequency firing (tetanic activity) within its constituent neurons. This synchronized synaptic activity triggers a cascade of biochemical changes, including NMDAR-induced calcium transients, leading to the upregulation of AMPA receptors and synaptic growth, thereby strengthening the connections within this circuit \citep{cite14}. This reverberating system is hypothesized to sustain excitation, bridging temporal gaps between stimulus and response, and serves as a fundamental ``element of thought.'' The formation of a memory engram, characterized by the strengthening of synaptic connections among selected neurons during encoding, is entirely consistent with Hebb's postulate \citep{cite13}.

Experimental evidence also strongly supports Hebb's principle. For example, \textit{in vitro} hybrid systems combining digital light processing (DLP) with optogenetics have been used to synchronize the firing patterns of two-neuron modules. These experiments demonstrate that synaptic strengthening predominantly occurs on dendrites between the simultaneously illuminated neurons, reflecting their coordinated activity. This strengthening can even be directed towards specific dendritic branches, leading to a patterned distribution of potentiated synapses, suggesting that Hebbian plasticity is not merely about increasing individual synaptic weights but actively sculpts the \textit{connectivity pattern} and even the \textit{topology} of the neuronal network \citep{cite13}. Synchronized activity, governed by Hebbian rules, drives the emergence of specific, reinforced network modules that constitute the physical engrams of memory.

\subsection{Mathematical Formulations of Hebbian Learning Rules}

The core principle of Hebbian plasticity can be mathematically formulated in various ways, ranging from simple correlational rules to more complex models incorporating homeostatic regulation.

A basic Hebbian learning rule for the change in synaptic weight ($\omega_{ij}$) between a presynaptic neuron $j$ (activity $u_j$) and a postsynaptic neuron $i$ (activity $v_i$) can be expressed as:
\begin{equation}
\frac{d\omega_{ij}}{dt} = \mu u_j v_i
\end{equation}
Here, $\mu$ represents a constant learning rate. This equation implies that if both neurons are active simultaneously, the strength of their connection increases.

However, simple Hebbian rules can lead to uncontrolled synaptic potentiation, potentially destabilizing neural networks. Biological systems overcome this through homeostatic mechanisms, such as synaptic scaling, which regulate overall synaptic strength. A more complex rule, known as \textbf{Synaptic Plasticity and Synaptic Scaling (SPaSS)}, combines Hebbian plasticity with a slower, homeostatic scaling rule \citep{cite17}. This combined rule ensures global stability and can be expressed as:

\begin{equation}
\frac{d\omega_{ij}}{dt} = \mu u_j v_i + \gamma(v_T - v_i)\omega_{ij}^2
\end{equation}

In this equation, $v_T$ is a target postsynaptic activity, and $\gamma$ is a rate factor that limits the scaling effect to a slower timescale than conventional plasticity. This can be further rewritten by defining $\kappa = \mu/\gamma$:

\vspace{-0.35cm}
\begin{equation}
\dot{\omega}_{ij} = \mu \left( u_j v_i + \kappa^{-1}(v_T - v_i)\omega_{ij}^2 \right)
\end{equation}

\vspace{-0.2cm}
For linear neuron models where the postsynaptic activity $v_i$ is a linear sum of weighted inputs ($v_i = \sum_j \omega_{ij} u_j$), a stable excitatory fixed point for the synaptic weight can be derived from this rule:

\vspace{-0.3cm}
\begin{equation}
\omega^*_{ij} = \frac{v_T}{2u_j} + \sqrt{\frac{\kappa}{u_j} + \left(\frac{v_T}{2u_j}\right)^2}
\end{equation}

\vspace{-0.2cm}
Alternatively, in terms of the stable postsynaptic activity $v^*_i$:
\begin{equation}
v^*_i = \frac{v_T}{2} + \sqrt{\kappa u_j^3 + \left(\frac{v_T}{2}\right)^2}
\end{equation}

\vspace{-0.25cm}
These equations demonstrate how homeostatic regulation is integrated into Hebbian learning to maintain stable network dynamics. The inclusion of a synaptic scaling term, as seen in the SPaSS model, is crucial for achieving global stability and preventing runaway synaptic potentiation. This indicates that biological Hebbian learning likely operates within a tightly regulated homeostatic context, where mechanisms exist to prevent excessive strengthening and maintain functional network dynamics. Computational models aiming for biological realism or robust continual learning must therefore incorporate such homeostatic or regularization mechanisms alongside Hebbian learning to ensure stability and long-term memory formation.

In sparse associative memory models, the Hebbian rule is also used to define synaptic strengths ($J_{ij}$) based on stored patterns ($S^\mu_i$, where $\mu$ indexes the patterns) \citep{cite18}:

\vspace{-0.275cm}
\begin{equation}
J_{ij} = \frac{1}{N} \sum_{\mu=1}^{M} S^\mu_i S^\mu_j
\end{equation}

\vspace{-0.2cm}
Here, $N$ is the total number of neurons and $M$ is the number of patterns to be stored. This formulation captures the essence of patterns being embedded in the network's connectivity.

\vspace{-0.15cm}
\subsection{Spike-Timing-Dependent Plasticity}

\vspace{-0.15cm}
While classic Hebbian learning emphasizes correlated activity, \textbf{Spike-Timing-Dependent Plasticity (STDP)} refines this principle by incorporating the precise temporal relationship between presynaptic and postsynaptic spikes \citep{cite19}. In this temporally asymmetric rule, if a presynaptic spike consistently precedes a postsynaptic spike within a short time window, the synapse is strengthened via \textbf{long-term potentiation (LTP)}. Conversely, if the presynaptic spike follows the postsynaptic spike, the synapse is weakened through \textbf{long-term depression (LTD)}. In contrast, \textbf{anti-Hebbian STDP} reverses this pattern: presynaptic spikes that precede postsynaptic activity lead to LTD, while those that follow induce LTP.

Computational models of \textbf{spiking neural networks (SNNs)} have extensively utilized STDP rules to investigate how memories are encoded and how neuronal assemblies form \citep{cite19}. These models demonstrate that inhibitory neurons, subject to both Hebbian and anti-Hebbian STDP, are crucial for the emergence and long-term maintenance of modularity within neural networks. Specifically, Hebbian inhibitory subpopulations help control overall firing activity, while anti-Hebbian inhibitory neurons promote pattern selectivity. This combination facilitates the formation of robust feedback and feed-forward inhibition circuits, which are essential for memory consolidation.

The critical role of inhibitory STDP in shaping network structure and function extends beyond merely suppressing activity. It actively participates in memory encoding by promoting pattern selectivity and helping to disentangle potentially overlapping memories, ensuring that distinct memory traces can be formed and retrieved without interference. The diverse inhibitory plasticity rules are therefore essential for creating distinct, non-interfering engrams and maintaining the stability and precision of network dynamics. Computational models of memory must explicitly incorporate these varied forms of inhibitory plasticity, rather than focusing solely on excitatory weight updates, to achieve biologically plausible memory capacity, effective pattern separation, and robust learning in complex, high-dimensional environments.

\begin{figure}[ht]
\centering
\begin{tikzpicture}[
    neuron/.style={circle, draw=black, fill=white, minimum size=12pt},
    arrow/.style={->, thick},
    excite/.style={->, thick, color=RoyalBlue},
    inhibit/.style={->, thick, color=Crimson, dashed},
    axis/.style={->, thick},
    every path/.style={line width=0.8pt},
    scale=0.92, transform shape
]

\node[neuron, label=left:Presynaptic] (pre) at (0,0.5) {};
\node[neuron, label=right:Postsynaptic] (post) at (2,0.5) {};
\draw[excite] (pre) -- (post);

\node at (1,1.5) {\small Hebbian Plasticity};

\draw[->, thick, gray!60] (pre) ++(-0.2,-0.6) -- ++(0.2,0.6);
\draw[->, thick, gray!60] (post) ++(-0.2,-0.6) -- ++(0.2,0.6);

\node at (1,-0.8) {\scriptsize ``Fire together $\Rightarrow$ wire together''};

\begin{scope}[xshift=5cm, yshift=-3.5cm]
\draw[axis] (-2.5,0) -- (2.5,0) node[right] {$\Delta t$ (ms)};
\draw[axis] (0,-1.5) -- (0,1.5) node[above] {$\Delta w$};

\draw[thick, RoyalBlue, domain=-2.5:0, samples=100] plot(\x,{1.2*exp(\x)});
\draw[thick, Crimson, domain=0:2.5, samples=100, dashed] plot(\x,{-1.2*exp(-\x)});

\node at (-1.8,1.2) {\tiny LTP};
\node at (1.8,-1.2) {\tiny LTD};

\draw[dotted] (-1,0) -- (-1,1.2);
\draw[dotted] (1,0) -- (1,-1.2);

\node at (0,-1.7) {\scriptsize Spike timing: $t_{\text{post}} - t_{\text{pre}}$};

\node at (0,2.2) {\small STDP Curve};
\end{scope}

\begin{scope}[xshift=7.5cm, yshift=-1.1cm]
\draw[axis] (0,0) -- (3.5,0) node[right] {Ca$^{2+}$ Concentration};
\draw[axis] (0,0) -- (0,2.5);

\draw[dashed] (1.1,0) -- (1.1,2.1) node[above] {\tiny LTD threshold\quad};
\draw[dashed] (2.2,0) -- (2.2,2.1) node[above, align=right] {\tiny \quad\quad\quad\quad\quad LTP threshold};

\fill[Crimson!40, opacity=0.6] (1.1,0) rectangle (2.0,1.2);
\fill[RoyalBlue!50, opacity=0.6] (2.2,0) rectangle (3.1,2.1);

\node at (1.75,2.75) {\small Ca$^{2+}$-Dependent Plasticity};

\end{scope}

\end{tikzpicture}

\caption{Illustration of Hebbian plasticity and Spike-Timing-Dependent Plasticity. Left: Coactivation of pre- and postsynaptic neurons strengthens the synapse (Hebb's Law; i.e., ``fire together, wire together''). Center: STDP curve showing how the relative timing of spikes controls potentiation (LTP) or depression (LTD). Right: Calcium concentration-based thresholds underlying synaptic change --- moderate calcium leads to LTD; high calcium leads to LTP.} %
\label{fig:stdp-diagram}
\end{figure}
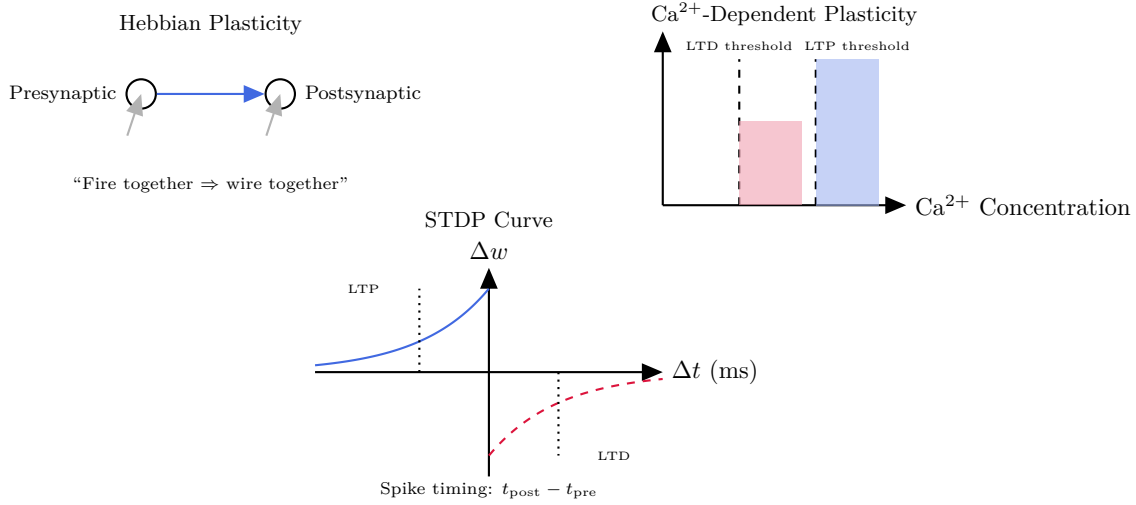

\vspace{-0.15cm}
\section{Sparsity Regularization in BNNs and ANNs}

\vspace{-0.15cm}
Sparsity, characterized by a small number of active components or connections, is a pervasive feature of \textbf{biological neural networks (BNNs)} and a critical principle in computational neuroscience for efficient memory systems.

\vspace{-0.15cm}
\subsection{The Biological Rationale for Sparse Coding in the Brain}

\vspace{-0.15cm}
In the brain, memory representations are often formed by sparse neuronal assemblies, particularly in regions like the hippocampus during learning \citep{cite3}. Engrams themselves are frequently described as ``sparse memory traces'' or ``sparsely distributed populations of neurons'' \citep{cite7, cite21}. Observations of sparse and selective neuronal activity during both memory encoding and recall are common across various brain areas, including the dentate gyrus and neocortex \citep{cite22}.

The prevalence of sparsity in biological memory systems is not coincidental; it offers significant advantages. A low level of neuronal activity is metabolically efficient, conserving the brain's considerable energy resources \citep{cite18}. Furthermore, sparse coding schemes demonstrably enhance both the memory capacity and information capacity of associative memories \citep{cite18, cite23}. For instance, theoretical models show that the memory capacity of a network can scale as $(N/\log(N))^2$ for very sparse activity, where $N$ is the number of neurons \citep{cite18}. This indicates that sparse coding allows for the error-free storage of a substantially larger number of patterns compared to dense coding.%

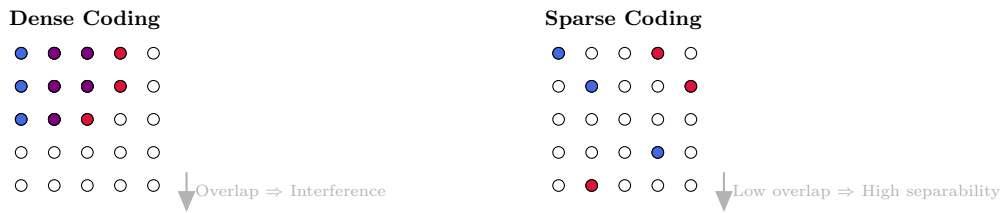
\begin{figure}[H]
\centering
\begin{tikzpicture}[
    neuron/.style={circle, draw=black, minimum size=5pt, inner sep=0pt},
    activeA/.style={circle, draw=black, fill=RoyalBlue, minimum size=5pt, inner sep=0pt},
    activeB/.style={circle, draw=black, fill=Crimson, minimum size=5pt, inner sep=0pt},
    overlap/.style={circle, draw=black, fill=Purple, minimum size=5pt, inner sep=0pt},
    gridnode/.style={minimum size=0pt, inner sep=0pt},
    labelstyle/.style={font=\small\bfseries},
    arrow/.style={->, thick, gray!60},
    scale=0.875, transform shape
]

\node[labelstyle] at (0.96,4.5) {Dense Coding};
\foreach \i in {0,...,4} {
  \foreach \j in {0,...,4} {
    \pgfmathtruncatemacro{\index}{\i*5+\j}
    \coordinate (d\index) at (\j*0.5,4 - \i*0.5);
    \node[neuron] at (d\index) {};
  }
}

\foreach \i in {0,1,2,5,6,7,10,11} {
  \node[activeA] at (d\i) {};
}
\foreach \i in {1,2,3,6,7,8,11,12} {
  \node[activeB] at (d\i) {};
}
\foreach \i in {1,2,6,7,11} {
  \node[overlap] at (d\i) {};
}

\draw[arrow] (2.5,2.2) -- ++(0,-0.6) node[midway, right] {\scriptsize Overlap $\Rightarrow$ Interference};

\begin{scope}[xshift=8.125cm]
\node[labelstyle] at (0.98,4.5) {Sparse Coding};
\foreach \i in {0,...,4} {
  \foreach \j in {0,...,4} {
    \pgfmathtruncatemacro{\index}{\i*5+\j}
    \coordinate (s\index) at (\j*0.5,4 - \i*0.5);
    \node[neuron] at (s\index) {};
  }
}

\foreach \i in {0,6,18} {
  \node[activeA] at (s\i) {};
}

\foreach \i in {3,9,21} {
  \node[activeB] at (s\i) {};
}

\draw[arrow] (2.5,2.2) -- ++(0,-0.6) node[midway, right] {\scriptsize Low overlap $\Rightarrow$ High separability};

\end{scope}

\end{tikzpicture}

\vspace{-0.2cm}
\caption{Comparison of dense (left) and sparse (right) memory coding. In dense coding, memories activate many overlapping neurons, increasing metabolic cost and interference. Sparse coding activates small, distinct neuronal subsets, improving efficiency and separability.}
\label{fig:sparse-vs-dense}
\end{figure}

The efficiency gains from sparsity stem from its ability to recode complex, correlated memories into compressed, more orthogonal representations \citep{cite24}. This orthogonalization makes memories more separable, facilitating their storage and reconstruction by reducing interference between different memory traces. Investigations into hippocampal autoencoder models suggest that optimal memory capacity is achieved when sparsity is \textit{weakly enforced}, rather than maximized. This implies a nuanced relationship, where the brain likely dynamically tunes its sparsity level based on factors like memory demand and the compressibility of incoming information. As such, this dynamic adjustment of sparsity represents a fundamental optimization principle that maximizes memory capacity and performance, adapting to the statistics of the environment.

\subsection{Computational Approaches to Sparsity Regularization}

In artificial neural networks, sparsity regularization methods are employed to mimic biological efficiency, reduce the number of non-zero parameters or activations, mitigate catastrophic forgetting, and improve model reliability \citep{cite7}. Regularization is typically achieved by adding a penalty term $R(\theta)$ to the original objective (loss) function $L(\theta)$ that the network aims to minimize. The optimization problem is generally formulated as:
\begin{equation}
\text{minimize } L(\theta) + \lambda R(\theta)
\end{equation}
where $\theta$ represents the network parameters (weights and biases), and $\lambda$ is a regularization factor controlling the strength of the penalty \citep{cite26, cite27}.

Several common computational approaches to enforce sparsity exist:

\begin{itemize}[leftmargin=*]
    \item \textbf{L0 Regularization:}
    L0 regularization directly penalizes the number of non-zero coefficients in the model. The penalty term is defined as the L0 ``norm'' (though it is not a true mathematical norm, lacking homogeneity):
    \begin{equation}
    R(\theta) = ||\theta||_0 = \sum_i \mathbb{I}(\theta_i \ne 0)
    \end{equation}
    where $\mathbb{I}(\cdot)$ is the indicator function, returning 1 if the condition is true and 0 otherwise \citep{cite28}. This method encourages strict sparsity, leading to models with only the most relevant variables. However, its non-convex and discrete nature makes it computationally expensive and often intractable for large-scale models \citep{cite28, cite29}. To overcome this, relaxed or differentiable approximations of L0 regularization are often used, for instance, by modeling the probability of a parameter being active using continuous distributions \citep{cite29}. A relaxed form of the L0 regularization loss can be expressed as:
    \begin{equation}
    L_0(\psi) = \sum_{j=1}^{|\xi|} (1 - P(d_j \le 0 | \psi))
    \end{equation}
    where $\psi$ indicates the probability of a gate (parameter) being active, and $P(d_j \le 0 | \psi)$ is the cumulative distribution function of a continuous random variable $d_j$ that represents the parameter's state \citep{cite29}.

    \item \textbf{L1 Regularization (Lasso):}
    L1 regularization adds a penalty proportional to the sum of the absolute values of the weights. The penalty term is the L1 norm:
    \begin{equation}
    R(\mathbf{w}) = ||\mathbf{w}||_1 = \sum_i |w_i|
    \end{equation}
    where $\mathbf{w}$ is the vector of weights \citep{cite27}. This method promotes sparsity by driving some weights to exactly zero, effectively performing feature selection \citep{cite27, cite30}. The regularized loss function becomes:
    \begin{equation}
    \tilde{\mathcal{L}}(\mathbf{w}) = \mathcal{L}(\mathbf{w}) + \alpha ||\mathbf{w}||_1
    \end{equation}
    where $\mathcal{L}(\mathbf{w})$ is the original loss function and $\alpha$ is the regularization strength \citep{cite27}. L1 regularization is convex and computationally efficient, making it widely used. It can control generalization error and sparsify input dimensions, potentially leading to near minimax optimal statistical risk \citep{cite30}.

    \item \textbf{KL-Divergence Penalty (for Sparse Autoencoders):}
    In sparse autoencoders, sparsity is enforced not on the weights directly, but on the \textit{activations} of the hidden units. This encourages only a limited number of hidden neurons to be active for any given input \citep{cite31}. The sparsity penalty term is typically incorporated into the autoencoder's overall cost function ($J_{sparse}$) using Kullback-Leibler (KL) divergence:
    \begin{equation}
    \beta \sum_{j=1}^{s_2} KL(\rho |
| \hat{\rho}_j)
    \end{equation}
    Here, $s_2$ is the number of hidden units, $\beta$ is a weight controlling the sparsity penalty, $\rho$ is the desired target sparsity level (a small value close to zero, e.g., 0.05), and $\hat{\rho}_j$ is the average activation of hidden unit $j$ across the training set \citep{cite31, cite32}. The KL-divergence between two Bernoulli distributions with means $\rho$ and $\hat{\rho}_j$ is given by:
    \begin{equation}
    KL(\rho |
| \hat{\rho}_j) = \rho \log\left(\frac{\rho}{\hat{\rho}_j}\right) + (1 - \rho) \log\left(\frac{1 - \rho}{1 - \hat{\rho}_j}\right)
    \end{equation}
    This term is minimized when $\hat{\rho}_j$ is equal to $\rho$, thereby pushing the average activation of each hidden unit towards the desired sparse level \citep{cite32}.

    \item \textbf{Engram Gating (Stochastic Engrams):}  
    A biologically inspired method of enforcing architectural sparsity involves the use of stochastically activated gating vectors to modulate neuron activity in \textbf{metaplastic binarized neural networks} (\textbf{mBNNs}; \citealp{cite7}). In this approach, input data is first encoded into a lower-dimensional latent space via a two-layer feedforward network. Each latent feature is then used to generate a binary gating vector through a Bernoulli sampling process, where the activation probability is determined by a sigmoid function applied to a real-valued hidden weight:
    \begin{equation}
    p = \sigma(z) = \frac{1}{1 + e^{-W_h}}, \quad W_s \sim \text{Bernoulli}(p)
    \end{equation}
    Here, $W_s$ represents the binary gate, controlling whether each neuron is active (1) or inactive (0) for a given task. This stochastically sampled gating vector is applied element-wise to the hidden layer, effectively selecting a sparse subnetwork --- or ``engram'' --- for learning. By dynamically allocating distinct subsets of neurons to different tasks, alongside the robustness of probabilistic memory traces, this mechanism mitigates catastrophic forgetting and promotes continual learning.

\end{itemize}

\vspace{0.5cm}
\begin{figure}[ht]
\centering
\begin{tikzpicture}[
    neuron/.style={circle, draw=black, fill=white, minimum size=8pt},
    active/.style={circle, draw=black, fill=RoyalBlue, minimum size=8pt},
    task/.style={rectangle, draw=black, fill=Gray!20, minimum height=6pt, minimum width=25pt},
    gatebit/.style={rectangle, draw=black, fill=white, minimum height=8pt, minimum width=8pt},
    onbit/.style={rectangle, draw=black, fill=DarkOrange, minimum height=8pt, minimum width=8pt},
    connection/.style={->, thick},
    gated/.style={connection, color=RoyalBlue},
    skip/.style={connection, gray!40},
    scale=1.5
]

\node[align=center, font=\small\bfseries] at (0,3.5) {Input\\(Task Representation)};
\foreach \i in {0,...,3} {
  \node[task] (t\i) at (0,2.5 - \i*0.6) {};
}

\draw[->, thick] (0.8,1.7) -- node[midway, above] {\footnotesize Encoder} (2.2,1.7);

\node[align=center, font=\small\bfseries] at (2.5,3.5) {Gating Vector\\(Engram)};
\foreach \i/\val in {0/0,1/1,2/0,3/1,4/0} {
  \ifnum\val=1
    \node[onbit] (g\i) at (2.5,2.7 - \i*0.5) {};
  \else
    \node[gatebit] (g\i) at (2.5,2.7 - \i*0.5) {};
  \fi
}

\node[align=center, font=\small\bfseries] at (5.5,3.5) {Binarized Neural Network\\(BNN)};
\foreach \i in {0,...,4} {
  \node[neuron] (b\i) at (5.5,2.7 - \i*0.5) {};
}

\foreach \i/\bit in {0/0,1/1,2/0,3/1,4/0} {
  \ifnum\bit=1
    \draw[gated] (g\i) -- (b\i);
  \else
    \draw[skip] (g\i) -- (b\i);
  \fi
}

\foreach \i in {1,3} {
  \draw[gated] (b\i) -- ++(1.0,0);
}
\foreach \i in {0,2,4} {
  \draw[skip] (b\i) -- ++(1.0,0);
}

\node[align=center, font=\small] at (7.8,1.7) {Task-Specific\\Forward Pass};

\end{tikzpicture}
\caption{Illustration of engram gating in a metaplastic binarized neural network. A task-specific encoder projects a gating vector that selects a sparse ensemble of neurons (a stochastic “engram”). Only this subset participates in forward pass and learning, preventing interference with other tasks and mitigating catastrophic forgetting.}
\label{fig:engram-gating}
\end{figure}
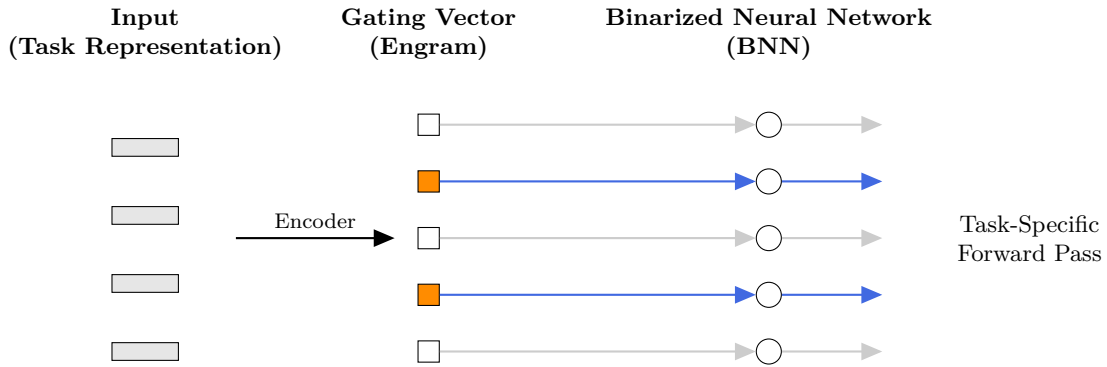

\clearpage
These different techniques illustrate a spectrum of approaches to achieving sparsity in ANNs. L0 and L1 regularization primarily affect the \textit{weights} of the network, driving them to zero, whereas the KL-divergence penalty in autoencoders targets the \textit{activations} of hidden units. Engram gating, on the other hand, represents an \textit{architectural} approach that dynamically allocates and manages neuronal ensembles. This progression in computational modeling reflects a growing understanding that sparsity in biological systems operates at multiple levels --- from individual synaptic strengths to the activity patterns of neuronal populations and the dynamic allocation of neural modules. The development of these diverse regularization techniques, each with distinct computational and biological implications, advances the ability to model and replicate the brain's efficient memory systems.

\vspace{-0.5cm}
\begin{table}[h!]
\centering
\caption{Comparison of Sparsity Regularization Methods}

\vspace{-0.15cm}
{\footnotesize\renewcommand{\arraystretch}{1.25}%
\begin{longtable}{|>{\raggedright\arraybackslash\scriptsize}p{2.5cm}|>{\raggedright\arraybackslash}p{3.5cm}|>{\raggedright\arraybackslash}p{2.75cm}|>{\raggedright\arraybackslash}p{2.5cm}|>{\raggedright\arraybackslash}p{3.25cm}|}
\hline
\centering\arraybackslash\footnotesize\textbf{Method} & 
\centering\arraybackslash\textbf{Mathematical Form (Penalty Term)} & 
\centering\arraybackslash\textbf{Effect on Network} & 
\centering\arraybackslash\textbf{Computational Properties} & 
\centering\arraybackslash\textbf{Relevance to Engrams/Memory} \\
\hline

\textbf{L0 Regularization} & $R(\theta) = ||\theta||_0 = \sum_i \mathbb{I}(\theta_i \ne 0)$ & Directly forces parameters to zero, leading to strict feature selection. & Non-convex, computationally intractable for large models; often approximated or relaxed. & Directly models the concept of ``few active neurons'' or connections in biological engrams. \\
\hline
\textbf{L1 Regularization (Lasso)} & $R(\theta) = ||\theta||_1 = \sum_i |\theta_i|$ & Shrinks weights, driving some to exactly zero; promotes sparsity. & Convex, computationally efficient; can be solved with gradient descent. & Induces sparse connectivity patterns, useful for memory compression and efficiency, mirroring biological mechanisms. \\
\hline
\textbf{KL-Divergence Penalty \quad\quad (for activations)} & $R(\hat{\rho}, \rho) = \rho \log\left(\frac{\rho}{\hat{\rho}_j}\right) + (1 - \rho) \log\left(\frac{1 - \rho}{1 - \hat{\rho}_j}\right)$ & Forces the \textit{average activation} of hidden units to a desired low value, ensuring sparse activity. & Differentiable, integrable into backpropagation; requires computing average activations. & Directly models sparse activity patterns observed in engram neurons, crucial for efficient encoding and pattern separation. \\
\hline
\textbf{Engram Gating (Stochastic Engrams)} & 
$p = \sigma(z) = \frac{1}{1 + e^{-W_h}}, \quad W_s \sim \text{Bernoulli}(p)$ & 
Applies binary gating vectors to hidden neurons to select sparse subnetworks per task; only gated neurons participate in learning. & 
Efficient for continual learning; combines binarized neural networks with stochastic sampling from low-dimensional latent features. & 
Implements context-specific engram formation and task-dependent memory routing, echoing biological principles of sparse ensemble allocation. \\
\hline
\end{longtable}
}
\end{table}

\vspace{-0.75cm}
\subsection{Advantages of Sparsity: Memory Capacity and Efficiency}

\vspace{-0.25cm}
The adoption of sparsity in both biological and artificial neural networks confers significant advantages, particularly concerning memory capacity and overall computational efficiency. Sparse coding schemes substantially enhance memory capacity in associative memories, allowing for the error-free storage of a large number of patterns \citep{cite18, cite23}. This increased capacity is partly due to sparse representations making memories more separable, thereby facilitating their storage and reconstruction by reducing interference \citep{cite24}.

Beyond capacity, a low level of neuronal activity, characteristic of sparsity, is metabolically advantageous for the brain, conserving energy resources \citep{cite18}. In computational models, sparse autoencoders demonstrate the ability to learn efficient representations with fewer active neurons, leading to reduced computational costs and more interpretable features \citep{cite31}.

However, the relationship between sparsity and memory performance is not always linear. Studies on sparse autoencoder models indicate that optimal memory capacity is achieved when sparsity is \textit{weakly enforced}, rather than maximized \citep{cite24}. This suggests a nuanced trade-off: while sparsity generally increases capacity, excessive sparsity might hinder the richness of representation necessary for complex memories, whereas insufficient sparsity reduces efficiency. Biological systems thus likely dynamically tune their sparsity level, finding an optimal balance that maximizes memory capacity and performance based on the complexity and compressibility of the input data.

\clearpage
\section{Computational Models: Plasticity and Sparsity}

Computational neuroscience is crucial in synthesizing biological observations into testable models, exploring how Hebbian plasticity and sparsity principles interact to form and maintain engrams.

\subsection{Engram Formation with Hebbian and Sparsity Constraints}

Computational models of spiking neurons are designed to investigate how memories are imprinted into the neuronal architecture, aiming to reproduce the formation of neuronal assemblies (engrams) mediated by synaptic plasticity \citep{cite19}. These models face inherent challenges, including ensuring the correct formation of neural assemblies, their long-term stability, and the overall stability of the resulting network dynamics. Advanced models incorporate both Hebbian and anti-Hebbian STDP, particularly involving inhibitory neurons. Such models can achieve stable modularity and pattern selectivity, which are crucial for effective memory consolidation. The interplay between Hebbian and anti-Hebbian inhibitory plasticity also allows for the formation of robust feedback and feed-forward inhibition circuits that control memory consolidation and ensure distinct memory representations.

\begin{figure}[H]
    \centering
    \includegraphics[width=0.67\linewidth]{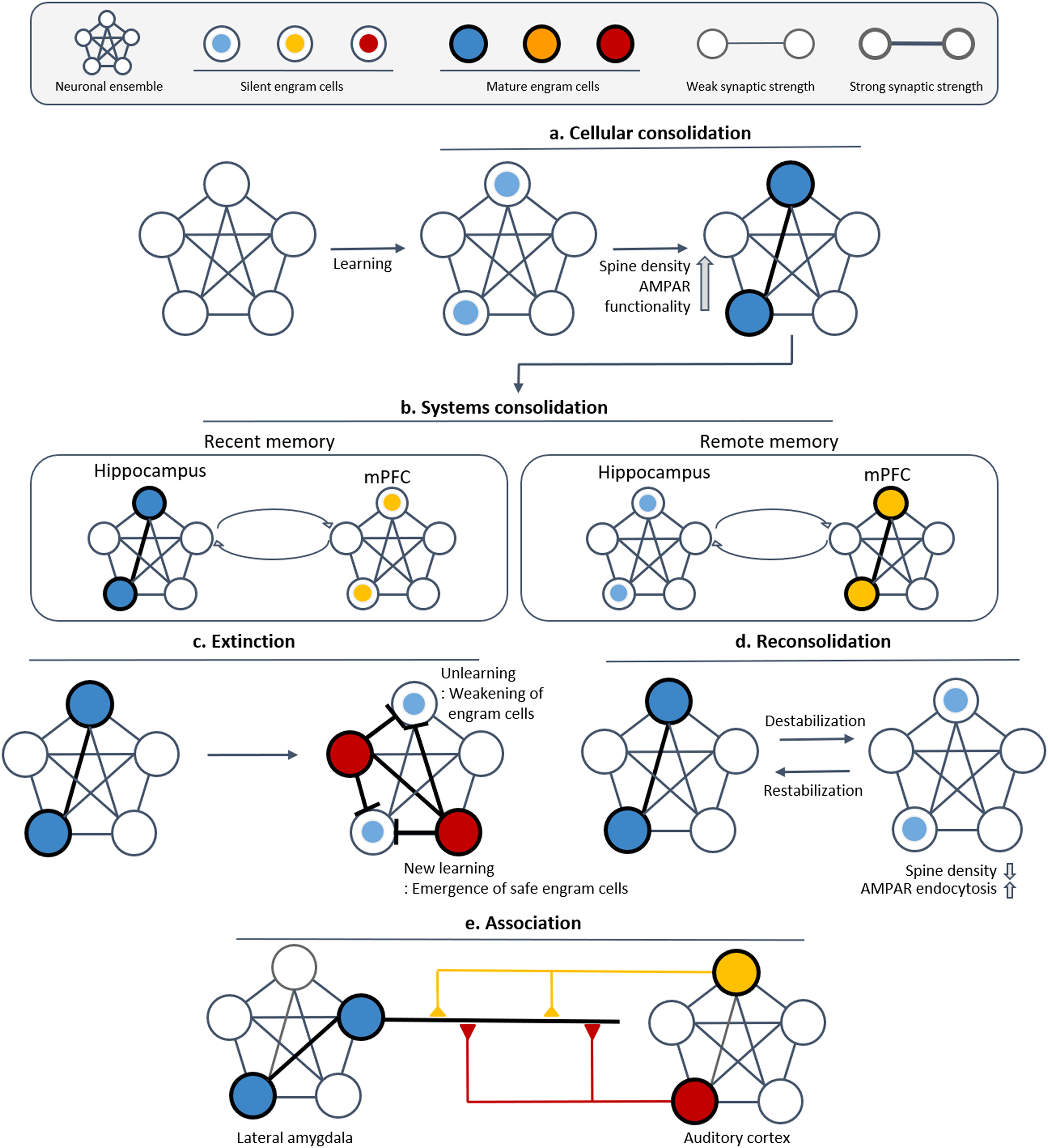}
    \caption{Phases of engram formation, transformation, and modulation (memory retrieval; adapted from \citealp{cite38}). (a) Cellular consolidation strengthens synapses and increases spine density and AMPAR functionality in engram cells. (b) Systems consolidation shifts memory dependence from the hippocampus to the medial prefrontal cortex (mPFC). (c) Extinction weakens the original engram and induces a new inhibitory “safety” engram. (d) Reconsolidation involves transient destabilization and restabilization of engram synapses following retrieval. (e) Associative learning links overlapping engram ensembles across regions, such as the amygdala and auditory cortex, via distinct synaptic pathways.}
    \label{fig:enter-label}
\end{figure}

Beyond synaptic weight changes, structural plasticity, where neurons dynamically grow and prune synaptic elements based on their activity, has also been shown to form ``silent'' memory engrams and implicitly model Hebbian learning \citep{cite33}. This mechanism allows for the simultaneous formation of multiple, non-interfering engrams, drawing inspiration from the columnar organization of the human cortex, where each column can represent a feature independently. As well, this suggests that the physical rewiring of connections (structural changes) is as important as, or may even precede, synaptic strength modulation, particularly for ensuring the distinctness and non-interference of stored memories. Structural plasticity contributes to the physical isolation and organization of engrams, preventing catastrophic interference between a large number of stored memories.

\vspace{-0.15cm}
\subsection{Sparse Distributed Memory Models and Associative Recall}

\vspace{-0.15cm}
\textbf{Sparse Distributed Memory (SDM)}, a mathematical model of human long-term memory introduced by Pentti Kanerva, provides a compelling framework for understanding associative recall and memory storage in high-dimensional spaces \citep{cite34}. SDM exhibits behaviors reminiscent of human memory, such as rapid recognition and the discovery of novel connections between seemingly unrelated ideas. 
In SDM, memories (patterns) are stored in a vast, high-dimensional binary space but are distributed across a smaller set of physical ``hard locations''. Memory retrieval in SDM relies on content-addressability, where a query vector is matched to stored patterns based on Hamming distance. For writing, if an input bit is 1, corresponding counters in selected hard locations are incremented; if 0, they are decremented. Reading involves summing the contents of selected locations column-wise and applying a threshold to reconstruct the memory. The model can also be conceptualized as a content-addressable memory or a three-layer feedforward neural network, as it strongly resembles the construction and behavior of the Perceptron model.

Recent work has shown that modified \textbf{Multi-Layer Perceptrons (MLPs)} incorporating features derived from SDM, such as a Top-K activation function (which keeps only the $k$ most excited neurons active) and L2 normalization, can achieve robust continual learning and effectively avoid catastrophic forgetting \citep{cite35}. The Top-K activation function in these models causes the most activated neurons to specialize towards specific inputs, leading to the formation of specialized subnetworks. Simultaneously, L2 normalization ensures that all neurons participate democratically in learning. SDM's inherent ability to avoid catastrophic forgetting, a major challenge in ANNs, stems from its core components that directly promote sparsity and democratic participation, echoing principles observed in biological engrams. The architectural and activation-based sparsity mechanisms inherent in SDM (e.g., Top-K activation) directly contribute to its ability to form distinct, non-interfering memory representations, thereby mitigating the problem of catastrophic forgetting. This makes SDM and its variants a promising direction for developing ANNs capable of lifelong learning, a key characteristic of biological memory.

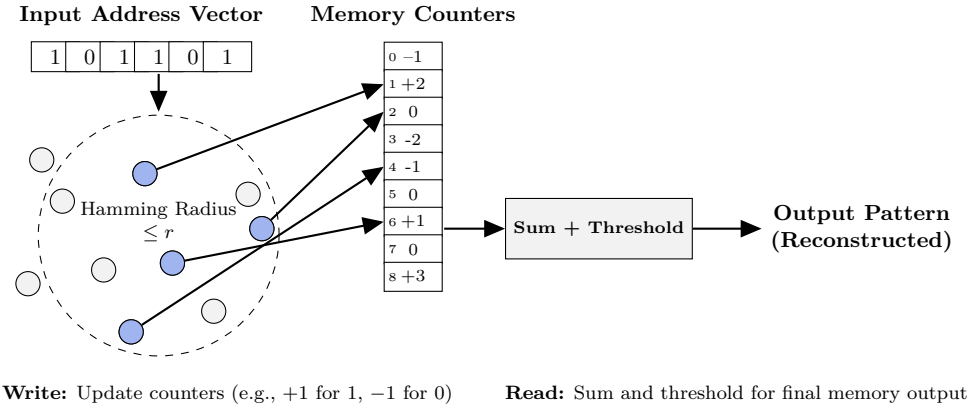
\begin{figure}[H]
\centering
\begin{tikzpicture}[
    addrbit/.style={rectangle, draw=black, fill=white, minimum width=20pt, minimum height=10pt, font=\small},
    hardloc/.style={circle, draw=black, fill=Gray!10, minimum size=10pt},
    selected/.style={circle, draw=black, fill=RoyalBlue!50, minimum size=10pt},
    counterrow/.style={rectangle, draw=black, fill=white, minimum width=24pt, minimum height=12pt, font=\scriptsize},
    sumbox/.style={rectangle, draw=black, fill=Gray!10, minimum width=60pt, minimum height=25pt, font=\scriptsize\bfseries},
    arrow/.style={->, thick},
    lab/.style={font=\scriptsize},
    scale=0.91, transform shape
]

\node at (1.25,4.7) {\small \textbf{Input Address Vector}};
\foreach \i/\b in {0/1,1/0,2/1,3/1,4/0,5/1} {
  \node[addrbit] (a\i) at (\i*0.5,4.1) {\b};
}

\draw[arrow] (1.5,3.85) -- (1.5,3.3);

\draw[dashed] (1.5,1.5) circle (1.75cm);
\node[align=center, font=\footnotesize] at (1.5,1.7) {Hamming Radius\\$\le r$};

\foreach \i/\x/\y/\bin in {
  0/0.1/2.0/011010,
  1/1.3/2.4/101011,
  2/3.0/1.6/101010,
  3/2.3/0.4/001111,
  4/1.1/0.1/111001,
  5/-0.4/0.8/110010,
  6/1.7/1.1/101110,
  7/2.8/2.1/100101,
  8/-0.2/2.6/011110,
  9/0.7/1.0/110101
} {
  \node[hardloc] (h\i) at (\x,\y) {};
}

\foreach \i in {1,2,4,6} {
  \node[selected] at (h\i) {};
}

\node at (5.2,4.7) {\small \textbf{Memory Counters}};
\foreach \j/\val in {0/–1,1/+2,2/0,3/-2,4/-1,5/0,6/+1,7/0,8/+3} {
  \node[counterrow] (c\j) at (5.2,4.1 - \j*0.4) {\val};
  \node[lab] at (4.9,4.1 - \j*0.4) {\tiny \j};
}

\foreach \i/\c in {1/1,2/2,4/4,6/6} {
  \draw[arrow] (h\i) -- (c\c.west);
}

\node[sumbox] (sum) at (7.9,1.6) {Sum + Threshold};

\draw[arrow] (5.65,1.6) -- (sum.west);

\draw[arrow] (sum.east) -- ++(1.0,0) node[right, align=center, font=\small\bfseries] {Output Pattern\\(Reconstructed)};

\node at (2.5,-0.8) {\footnotesize \textbf{Write:} Update counters (e.g., $+1$ for 1, $-1$ for 0)};
\node at (9.9,-0.8) {\footnotesize \textbf{Read:} Sum and threshold for final memory output};

\end{tikzpicture}

\vspace{-0.15cm}
\caption{Sparse Distributed Memory: During \textbf{write}, a binary input address activates nearby hard locations within a Hamming radius, and their counters are updated (+1 for a 1, -1 for a 0 in the input pattern). During \textbf{read}, those same hard locations are accessed again, their counters are summed column-wise, and a threshold is applied to reconstruct the stored memory. This enables fault-tolerant, high-capacity, content-addressable memory.}
\label{fig:sdm-column-counters}
\end{figure}

\subsection{Spiking Neural Network Models for Engram Dynamics}

\vspace{-0.15cm}
SNNs represent a class of computational models that aim for higher biological realism by simulating the discrete firing events (spikes) of neurons, rather than continuous firing rates \citep{cite19}. SNNs are well-suited for incorporating detailed Hebbian and anti-Hebbian STDP rules to investigate the intricate processes of memory encoding and neuronal assembly formation. %
Models based on SNNs have demonstrated that specific inhibitory sub-populations are essential for achieving stable modularity and pattern selectivity in neural networks. After learning, these networks can settle into asynchronous irregular resting-states, reflecting a balanced and efficient network dynamic.%

\vspace{-0.1cm}
Despite their promise, SNN models face significant challenges in fully capturing the complexity of biological engram dynamics. The ``incredible complexity of the neuronal hardware'' in the brain, including factors like neuromodulation, neuronal and synaptic adaptation across various spatial and temporal scales, the vast diversity of neuron types, and the crucial role of glial cells, remains largely unintegrated into current SNN theories \citep{cite6}. While current SNNs successfully demonstrate fundamental principles, fully capturing engram dynamics necessitates a more comprehensive integration of these biological details. This presents a tension between computational traceability and biological fidelity; hence, progress in this area relies on progressively incorporating more biological detail into SNNs while simultaneously developing more efficient simulation techniques. %

\vspace{-0.25cm}
\section{Challenges and Future Directions}

\vspace{-0.25cm}
Despite significant advancements, engram research --- both biological and computational --- faces several formidable challenges that define its future trajectory.

\vspace{-0.25cm}
\subsection{Methodological and Theoretical Hurdles in Engram Research}

\vspace{-0.23cm}
The precise identification of memory engrams remains a substantial methodological obstacle in neuroscience \citep{cite36, cite37}. Despite considerable progress in labeling and manipulation methodologies, engrams have not been ``clearly identified,'' and their exact appearance or characteristics within the brain are still unclear \citep{cite37}. A critical issue is an ``epistemic bias'' in engram neuroscience, which tends to prioritize characterizing biological changes while neglecting the concurrent development of robust theoretical frameworks. The field is described as ``under-theorised,'' implying that without a comprehensive theory of ``what engrams are, what they do, and the wider computational processes they fit into, it may never be definitively known when they have been found.''

\vspace{-0.1cm}
The inherent properties of memory itself complicate the very concept of an engram. Memory storage models often propose distributed networks where information is represented by overlapping activity patterns across multiple nodes \citep{cite36}. Furthermore, memories are not static, accurate recordings of past events; they are prone to errors, distortions, and are actively reconstructed and reconsolidated over time. This malleability raises questions about the stability and distinctiveness of engrams as stable representations. It remains unclear what specific information content an engram holds and how this information persists over long durations \citep{cite37}.

\vspace{-0.1cm}
Experimental limitations also persist, including the temporal resolution of current cell-tagging protocols, variations in immediate early gene function, challenges in controlling for off-target effects of genetic manipulations, and the difficulty in precisely distinguishing engram from non-engram cells \citep{cite37}. From a computational perspective, current neural network models applied to engrams still have a relatively limited repertoire of cognitive behaviors they can explain \citep{cite6}. A significant challenge lies in integrating the full richness of the brain's neuronal hardware, including neuromodulation, various forms of neuronal and synaptic adaptation, the incredible diversity of neuron types, and the complex roles of glial cells, into theoretical network models.

\vspace{-0.1cm}
The pervasive ``theory gap'' in engram neuroscience is a critical impediment. The lack of a unifying theoretical framework hinders the comprehensive interpretation of complex experimental data and limits the explanatory power of current engram research. Computational neuroscience is uniquely positioned to address this by developing comprehensive, testable computational theories of engram function, necessitating greater collaboration between experimentalists and theoreticians to ensure that models are both biologically constrained by empirical data and capable of generating novel, empirically verifiable predictions.

\subsection{Therapeutic Implications for Memory Disorders}

\vspace{-0.2cm}
The burgeoning understanding of engrams and their dynamics holds profound therapeutic implications for a range of memory disorders. A deeper understanding of engram formation, consolidation, and activation is considered critical for developing effective treatments for conditions such as Alzheimer's disease \citep{cite4}. Research suggests that memory dysfunction in disorders like Alzheimer's may stem from pathologies in the early window after memory formation, specifically when engrams are undergoing crucial changes and stabilization. Studies are actively investigating mouse models of early Alzheimer's to determine if engrams form but fail to stabilize correctly \citep{cite4, cite20}. This research aims to identify specific genes that are altered during engram refinement and consolidation, which could then serve as targets for genetic or pharmacological modulation to improve memory performance \citep{cite20}.

Beyond neurodegenerative diseases, engram research offers valuable insights into conditions like Post-Traumatic Stress Disorder (PTSD; \citealp{cite21}). In PTSD, maladaptive or traumatic memories persist and are readily reactivated. The emerging understanding of natural forgetting as an adaptive form of engram plasticity, involving a reversible suppression of engram ensembles, opens new avenues for therapeutic intervention \citep{cite15}. If forgetting can be modulated as an active process, it might be possible to selectively suppress or diminish the accessibility of pathological memories, thereby alleviating symptoms of conditions where maladaptive memories are a core feature.

The direct link between engram dynamics and memory disorders suggests that manipulating engrams could offer novel therapeutic avenues. This indicates a shift towards ``precision medicine'' in memory disorders, moving beyond symptomatic treatment to targeting the fundamental neural substrates of memory. A deeper understanding of engram pathology can directly inform the development of targeted genetic or pharmacological interventions, potentially leading to more effective and personalized treatments. Computational models can also play a crucial role in this translational effort by simulating different disease states and intervention strategies at the engram level, accelerating the discovery and optimization of effective therapies.

\vspace{-0.3cm}
\section{Conclusion}

\vspace{-0.3cm}
The concept of the engram, initially a theoretical abstraction by Richard Semon, has evolved into a tangible biological entity representing the physical trace of memory in the brain. Modern neuroscience, armed with advanced technologies like optogenetics and immediate early gene labeling, has provided compelling evidence for the existence and dynamic nature of engram neurons and their ensembles. These studies reveal that engram formation involves intricate cellular and synaptic changes, including the strengthening of connections via Hebbian plasticity and the sculpting of network architecture through both excitatory and inhibitory plasticity mechanisms.

The integration of Hebbian plasticity and sparsity within computational models is crucial for advancing the understanding of memory. These models, ranging from abstract associative memories to biologically realistic spiking neural networks, provide frameworks for investigating how memories are encoded, consolidated, and retrieved. Sparsity thus emerges as a critical organizing principle in both biological and computational memory systems. In the brain, sparse coding enhances memory capacity, promotes pattern separation, and improves metabolic efficiency. Computational neuroscience has mirrored these biological strategies through regularization techniques (e.g., L0, L1, KL-divergence), which support sparse representations in artificial networks. Models like Sparse Distributed Memory further demonstrate how architectural sparsity can enable robust continual learning and mitigate catastrophic forgetting.

Taken together, these insights support a computational theory of engram encoding and retrieval in which memory arises from the interaction of sparse high-dimensional representations, temporally sensitive synaptic plasticity, and structural refinement over time. In this framework, engram formation is gated by activity-dependent plasticity, stabilized through consolidation, and retrieved via partial pattern reactivation. While significant methodological and theoretical hurdles remain, particularly in characterizing engram content and integrating the brain’s complex neuronal hardware, the growing synergy between neurobiology and computational modeling offers a promising path toward unified models of memory and cognition. This interdisciplinary foundation not only deepens our understanding of memory but also guides the development of therapeutic strategies for memory-related disorders.

\setlength{\bibsep}{13pt plus 0.3ex}
\bibliographystyle{tmlr}
\bibliography{engram_references}

\end{document}